# A Learning-Based Tune-Free Control Framework for Large Scale Autonomous Driving System Deployment

Yu Wang[1], Shu Jiang[1], Weiman Lin, Yu Cao, Longtao Lin, Jiangtao Hu, Jinghao Miao and Qi Luo[2]

*Abstract*— This paper presents the design of a tune-free (human-out-of-the-loop parameter tuning) control framework, aiming at accelerating large scale autonomous driving system deployed on various vehicles and driving environments. The framework consists of three machine-learning-based procedures, which jointly automate the control parameter tuning for autonomous driving, including: a learning-based dynamic modeling procedure, to enable the control-in-the-loop simulation with highly accurate vehicle dynamics for parameter tuning; a learning-based open-loop mapping procedure, to solve the feedforward control parameters tuning; and more significantly, a Bayesian-optimization-based closed-loop parameter tuning procedure, to automatically tune feedback control (PID, LQR, MRAC, MPC, etc.) parameters in simulation environment. The paper shows an improvement in control performance with a significant increase in parameter tuning efficiency, in both simulation and road tests. This framework has been validated on different vehicles in US and China.

## I. INTRODUCTION

### A. Motivation

Scalability is critical in autonomous driving system deployment and commercialization. Apollo platform [1] aims to enable autonomous driving in diverse scenarios [2] across nations and vehicles [3]. However, for each different scenario or vehicle, control parameters need to be re-tuned for performance purpose. As a consequence, parameter tuning becomes one of the biggest challenges in large-scale deployment. The generalized parameters not only refer to control gains, but also include continuous/discrete decision parameters, graph search/optimization coefficients and machine learning hyper-parameters. To resolve the issue, we propose an automatic end-to-end parameter tuning framework, that helps to scale up autonomous driving system operations on different vehicles with enhanced efficiency and reduced cost.

### B. Related Works

Automatic parameter tuning for robotic systems has been a long time industrial need and research interest. Early works were usually limited to simple PID controllers utilizing Integral of Time Absolute Error (ITAE) optimization [4], Ziegler-Nichols method [5], or simulated annealing [6]. [7] presented a new design of Predictive Functional Control (PFC) to achieve consistent parameter tuning results, but lacked generalization for different controllers. In [8], an iterative parameter tuning method was presented, but it was only limited to feedforward controllers. Other works, such as evolutionary algorithms [9] for automatic parameter tuning in robotics [10] [11] can extend their methods to general control systems but were known for data inefficiency.

Recent works had alternative approach of using Reinforcement Learning (RL) for automatic parameter tuning [12] or extended neural network architecture search [13]. The applications of this approach were expanded to all the modules of robotic systems (from perception motion tracking [14] to control). While the RL approach usually outperforms previous methods, it has significant drawbacks of poor scalability (since RL algorithm is coupled with its specific application) and high computational cost. [15] formulated the parameter tuning of a collective pitch controller as a multi-objective optimization problem and solved it with game theory.

Bayesian-based optimization method, on the other hand, has the characteristics like: *high data efficiency*, *black box*, *support for both continuous and discrete parameters*, thus has been largely discussed and applied in both academia [16] and industries [17] [18] as cloud package service for different applications. In [19], a Bayesian optimization algorithm was proposed to tune both the feedforward and joint PID gains of a robot manipulator controller. [20] presented a Bayesian-based calibration strategy to tune the Model Predictive Control (MPC) cost with little additional experimental effort. However, rare work has been done in building upon and applying this technology to autonomous vehicle on-board parameter tuning, with systematic autonomous driving control performance profiling.

### C. Contributions

In this paper, a learning-based framework that automatically tunes the autonomous driving onboard parameters is presented, which makes the control module completely tune-free (human-out-of-the-loop tuning). Our contributions are summarized as:

1) **A complete control tune-free framework**: This includes three major parts as shown in Fig. 1.
   a) *Learning-based dynamic model*: the dynamic model of ego vehicle, generated by supervised machine learning, provides the plant model of the control-in-the-loop simulation and serves as the essential precondition of the feedback control parameter tuning in simulated environment;

[1] Equally contributed to this paper
[2] Corresponding author
All Authors are with Baidu USA LLC, 1195 Bordeaux Dr, Sunnyvale, CA 94089 luoqi06@baidu.com

b) *Learning-based open-loop mapping*: the 3-dimensional calibration table generated by supervised machine learning, provides the mapping between longitudinal command (acceleration), state (speed) and actuation (throttle/brake) and serves as the feedforward control parameter tuning;
c) *Bayesian-optimization-based feedback control auto-tuner*: the Bayesian-based machine learning tool, automates the parameter tuning for feedback controls including *PID, LQR, MRAC and MPC*, and thus makes the whole control module tune-free. This is the main-body of the framework.
2) **Real world application and open-accessibility**: Following extensive onboard and offboard testings, this system has been deployed to several fleet of self-driving vehicles of different types in both China and US. Apollo platform may provide the tune-free control as a public service as we have done for other services.

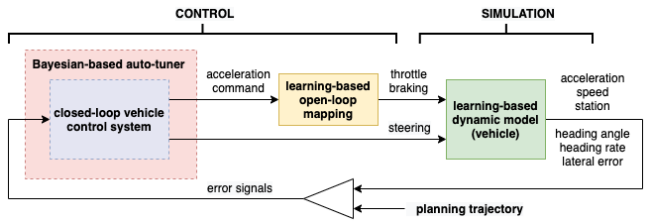

Fig. 1: Complete tune-free control framework

This paper is organized as: Section II introduces the tune-free control framework fundamentals; Section III elaborates the main body of tune-free control framework; Section IV presents the simulation and road-test results, validating the framework's efficiency and robustness.

## II. TUNE-FREE CONTROL FRAMEWORK FUNDAMENTAL

This section will introduce the learning-based dynamic model and open-loop mapping, which are two essential components making the closed-loop auto-tune possible.

### A. Learning-based Vehicle Dynamic Modeling for Simulation

The Apollo simulation platform supports the control-in-the-loop simulation, in which the rule-based or learning-based dynamic models are designed to ensure the ego vehicle's dynamic characteristics in real world can be truthfully reproduced in simulated environment, with thousands of virtual city-road and traffic scenarios.

Dynamic model bridges the gap between control command and vehicle dynamic behaviors in simulation environment. As shown in Fig. 2, dynamic model takes control command as input and publishes vehicle states (i.e., replaces the real-world chassis and localization module) in simulation.

The rule-based vehicle dynamic modeling method usually lacks generality due to the high cost on identifying lots of vehicle-by-vehicle varying parameters in the complex dynamic mechanism. We adopt a data driven machine-learning-based method, which requires much less single-vehicle developing cost and hence can be easily generalized. To better mimic the high phase lag plus time delay involved in the vehicle dynamics, our learning-based dynamic model employs a recurrent neural network model, Long Short-Term Memory (LSTM) by leveraging its advantage in processing long historical data sequence. The model design is summarized in Table I and visualized in Fig. 3a. The input layer of the learning-based dynamic model has five dimensions $[u_t, u_b, u_{st}, v_i, a_i]$, where $u_t, u_b, u_{st}$ represent throttle, brake and steering control command respectively. $v_i$ is ego vehicle's speed and $a_i$ is acceleration at current sample point. The output layer has two dimensions $\left[a_{i+1}, \dot{\theta}_{i+1}\right]$, where $a_{i+1}$ is the acceleration and $\dot{\theta}_{i+1}$ is the heading angle change rate for next sample point.

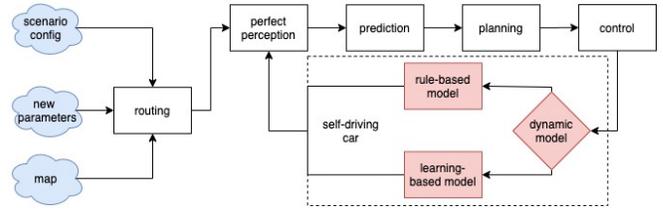

Fig. 2: Simulation-based autonomous driving system

This modeling procedure integrates with standard GUI-guided data-collection and model training processes, and hence can automatically generate single-vehicle dynamic model within hours, which is about 16 times faster than over-one-week manual work. Fig. 3b shows the higher model accuracy of LSTM-based modeling than the rule-based one and other learning models (such as MLP). More details about our learning-based dynamic modeling are presented in [3].

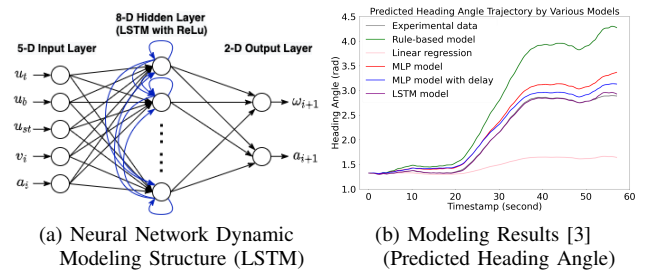

(a) Neural Network Dynamic Modeling Structure (LSTM)

(b) Modeling Results [3] (Predicted Heading Angle)

Fig. 3: Dynamic modeling model structure and results

### B. Learning-based Open-Loop Mapping

In the vehicle control system, to avid analyzing the complex powertrain dynamics and provide the accurate feedforward control, the drivetrain dynamics between control outputs, i.e. brake and throttle, and vehicle states, i.e. acceleration and deceleration, are quantified by a table, named calibration table. Parameter tuning of calibration table is the very first step of building an autonomous vehicle longitudinal control system (refer to Section III-A.1). We call this process as open-loop mapping. Our learning-based mapping is generated by a three-layer Multilayer Perceptron (MLP) model with actuation delays encoded, as shown in in Table I. MLP is a typical feedforward neural network structure, which is especially fit for the input/output mapping problem.

TABLE I: Supervised machine learning features

|  | **Dynamic Modeling** | **Open-loop Mapping** |
|---|---|---|
| Model Type | LSTM (with 20-step-long input sequence) | MLP |
| Neural Network Structure | a 5-D input layer, an 8-D hidden layer with ReLU activation function, and a 2-D output layer | a 2-D input layer, a hidden layer with Sigmoid activation function, and a 1-D output layer |
| Training Environment | 6 computing nodes, each has 4 i7-CPUs clocked at 2.6GHz | 6 computing nodes, each has 4 i7-CPUs clocked at 2.6GHz |

The learning model takes the throttle (or brake) command and current speed $[u_t(u_b), v_i]$ in the input layer and predicts the acceleration $[a_{i+1}]$ at the next sample at the output layer. The internal layer is a fully connected layer represented as:

$$sig(Z_l) = sig(\omega_1 u_t + \omega_2 v_i + \rho_l) \quad (1)$$

Where $sig(Z) = 1/(1+e^{-Z})$ is the Sigmoid function which captures non-linear relationship between inputs. $\omega_1, \omega_2$ are input weights and $\rho_l$ is the bias in the neural network. The dimension of the hidden layer $l = 1...N_l$ is the main model hyperparameter which is optimized via enumeration method.

The standard GUI-guided data-collection and cloud-based model training processes save the engineers from the manual control calibration workflow, and therefore reduces the data collection and training time from days to around two hours, which increases the time-efficiency by around 20 times. Fig. 4a-4b show the distribution of the collected raw data and trained open-loop mapping results, respectively.

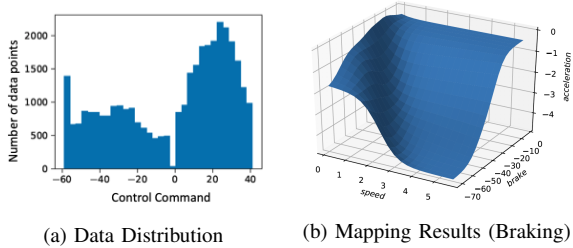

(a) Data Distribution  (b) Mapping Results (Braking)

Fig. 4: Open-loop mapping results

## III. TUNE-FREE CONTROL FRAMEWORK MAIN-BODY

As the main-body of the tune-free control framework, our closed-loop control system utilizes the Bayesian-based auto-tuner to realize parameter-tuning automation.

### A. Closed-Loop Control Design

This section focuses on the design of closed-loop control system used in both road tests and control-in-the-loop simulation. The closed-loop control algorithm processes the feedback control errors, i.e., the longitudinal station error $e_x$, speed error $e_v$, lateral error $e_y$ and heading error $e_\theta$, and generates the throttle-or-brake command $u_{tb}$ and steer command $u_{st}$. The control algorithm is visualized in a dual-loop closed-loop control architecture as shown in Fig. 5. The outer-loop control system addresses the vehicle longitudinal and lateral dynamic control problems; while the inner-loop control system compensates the by-wire actuation dynamic delay.

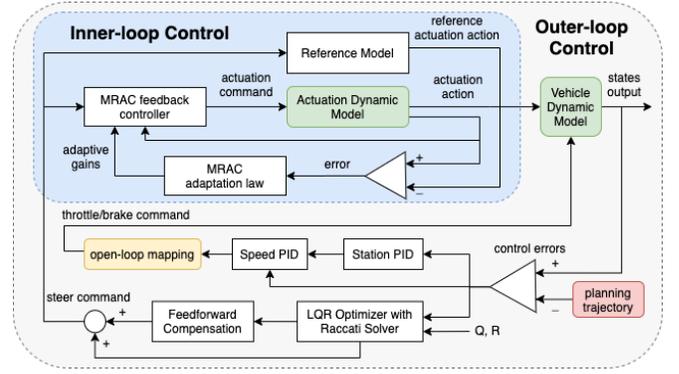

Fig. 5: Closed-loop control architecture

*1) Outer-Loop Controller:* The outer-loop controller tackles the high-dimension vehicle dynamics control problem by decoupling it into the longitudinal control and lateral control.

The longitudinal feedback controller cooperates with the open-loop mapping to manipulate the throttle/brake command $u_{tb}$ to address the powertrain and drive resistance dynamics. The open mapping plays a role in the feedforward control to help simplify the feedback control design. The longitudinal controller consists of the feedforward control $f_{map}()$ plus the cascaded station and speed PID controllers, given by:

$$\begin{aligned} a_{cmd} &= a_{ref} + K_{Pv}(e_v + K_{Px}e_x) + K_{Iv}\int(e_v + K_{Px}e_x) \\ u_{tb} &= f_{map}(a_{cmd}, v) \end{aligned} \quad (2)$$

where, $a_{cmd}$ is the acceleration command, $a_{ref}$ is the previewed acceleration reference predicted from planning trajectory. $K_{Pv}$, $K_{Iv}$, and $K_{Px}$ are the proportional and integral gains of the speed and station errors, respectively. The gain scheduler further splits $K_{Pv}$ and $K_{Iv}$ into high-speed gains ($K_{Pv,h}$, $K_{Iv,h}$) and low-speed gains ($K_{Pv,l}$, $K_{Iv,l}$).

For longitudinal control, there are five parameters selected for auto-tuning: ($K_{Pv,h}$, $K_{Iv,h}$, $K_{Pv,l}$, $K_{Iv,l}$ and $K_{Px}$ are selected for the auto-tuner.

The lateral controller employs the Linear-Quadratic-Regulator (LQR) control algorithm to manipulate the steer command $u_{st}$ to handle the lateral translation/rotation dynamics and tire-friction dynamics, which are described by the $4^{th}$-order nonlinear models $\dot{X}_y = A_y X_y + B_{y1}\delta + B_{y2}\dot{\theta}_{des}$. Here, the states $X_y = [e_y, \dot{e}_y, e_\theta, \dot{e}_\theta]^T$; the input $u_{st} = k_{st}\delta$, where $\delta$ is front-tire steering angle and $k_{st}$ is an angle-to-steer constant; the augmented term $\dot{\theta}_{des}$ is the desired heading angle rate. The details of matrix $A_y$, $B_{y1}$ and $B_{y2}$ are interpreted in [21].

The selection of the LQR control is attributed to its analytical solution and fast computational speed. It transforms the problem into a linear error-dynamics optimization control with feedback gains $K_y$, plus a nonlinear steady-state feedforward term $\delta_{ff}$ to eliminate the augmented dynamic terms, given by:

$$\delta_{cmd} = -K_y X_y + \delta_{ff}$$
$$\delta_{ff} = L_w \kappa + k_v v^2 \kappa + K_{y3}\theta_{e,ss} \quad (3)$$

where, linear feedback gains $K_y = [K_{y1}, K_{y2}, K_{y3}, K_{y4}]^T$, $L_w$ is the wheel-base length, $\kappa$ is the driving path curvature, $k_v$ is the vehicle understeer gradient, and $\theta_{e,ss}$ is the steady-state yaw-angle error calculated with the Final Value Theorem.

Substituting (3) into the original nonlinear model gives us a linear transformed system $\dot{X}_y = (A - B_{y1}K_y)X_y$. On this basis, $K_y$ are optimized by solving the Riccati equation with the LQR cost function $J$, given by:

$$J = \sum_{k_t=0}^{N_t} \left[ X_y(k_t)^T Q X_y(k_t) + \delta(k_t)^T R \delta(k_t) \right] \quad (4)$$

where, $k_t = 0...N_t$ denotes the LQR optimization horizon, $Q$ and $R$ are the states and input weighting matrix, respectively.

For lateral control, four parameters $Q = [Q_1, Q_2, Q_3, Q_4]^T$ dominate the controller and thus are selected for auto-tuning.

Alternatively, the outer-loop control can be realized by a Model-Predictive-Control (MPC) controller, for which no more details are provided due to the length limitation.

*2) Inner-Loop Controller:* The Inner-loop controller is targeted at fixing the actuation dynamics introduced by the by-wire steering mechanism, which is not included in the outer-loop control models due to lack of system parameters. To control this unknown-parameter dynamics, the inner-loop Model-Reference-Adaptive-Control (MRAC) introduces a "desired" steer actuation model as reference, and adapts the gains in real-time for re-building the steer command to compensate the delayed steering actuation dynamics. Our MRAC controller supports $n^{th}$-order model ($n \geq 1$); without losing the generalization, the real and reference actuation dynamics are given by $1^{st}$-order model as:

$$\dot{\delta} = \alpha\delta + \beta\delta_{mrac}, \quad \dot{\delta}_{ref} = \alpha_{ref}\delta_{ref} + \beta_{ref}\delta_{cmd} \quad (5)$$

where, $\delta$, $\delta_{ref}$ are the actual and reference angles, $\alpha$, $\beta$, $\alpha_{ref}$ and $\beta_{ref}$ are the actual and reference model constants, respectively. $\delta_{cmd}$ is the original steer command inherited from outer-loop controller, while $\delta_{mrac}$ is the post-MRAC steer command to produce better lateral path-tracking performance.

By saving the lengthy proof process with Lyapunov Stability Theorem, the adaptive control law $\delta_{mrac}$ is given by:

$$\delta_{mrac} = \hat{K}_\delta \delta + \hat{K}_u \delta_{cmd}$$
$$\dot{\hat{K}}_\delta = -p\gamma_\delta \delta(\delta - \delta_{ref}) \quad \dot{\hat{K}}_u = -p\gamma_u \delta_{cmd}(\delta - \delta_{ref}) \quad (6)$$

where, $\hat{K}_\delta$, $\hat{K}_u$ are the adaptive gains, $\gamma_\delta$, $\gamma_u$ are convergence-rate gains, for real and reference systems, respectively.

For MRAC control, the reference model time-constant $T_{ref}$ and additional adaption gain $p$ are the auto-tuned parameters.

### B. Closed-Loop Parameter Auto-tuner

This section introduces the auto-tuner and how it utilizes Bayesian-based optimization methods to tune control parameters via the iterative simulation tests.

First, we set up an input configuration file to initialize the auto-tuner, including: selecting a set of tunable parameters and their bounding numerical ranges, selecting simulation scenarios, and fixing the auto-tune iteration steps, etc.

As shown in Fig. 6, the auto-tuner includes three parts: 1.) A cloud-based simulation environment that runs autonomous driving full stack (from perception to control) and, on this basis, distributes tens of suggested optimal parameters in thousands of scenarios in parallel to generate running records/history under these parameters, as introduced in Section II-A. 2.) A performance profiling module that grades the simulation records generated from previous part. 3.) A Bayesian-based optimizer that efficiently suggests new trials of parameters based on previous grading scores. This finishes one iteration and sends back to simulation environment again for next iteration.

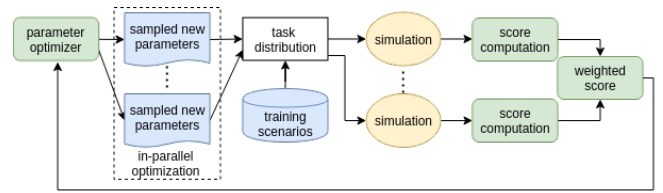

Fig. 6: Simulation-based autotuner flowchart

*1) Performance Profiling (tuner grader):* The tuner grader extracts information from the simulation records, one from each selected simulation scenarios, and evaluates the control performance by computing up to 41 profiling metrics. The profiling metrics cover four control-related aspects: 1.) control performance, 2.) driving safety, 3.) driving comfort and 4.) actuator efforts. Examples of important control profiling metrics are listed in Table II. The three lateral-error metrics evaluate the worst performance, statistically average (root-mean-square, RMS) performance, and statistically average performance at some "harsh" driving zones with high path curvature, respectively. The re-planning event metric is to penalize the occurrences of changing the control reference (planning trajectory) due to large control errors. All the other metrics (involved with station error, speed error, heading error, heading rate error, etc.) are presented in [22].

The combined evaluation score (or grade) $g$, which is the objective of the Bayesian optimization, is calculated by multiplying each single-metric score $s$ with its normalized weight coefficient $w$:

$$g_i = \sum_{j=1}^{N_s} \left\{ \left[ \sum_{k=1}^{N_m} s_{j,k}(h_i) w_k(h_i) \right] \frac{m_j}{\sum_{j=1}^{N_s} m_j} \right\} \quad (7)$$

where, $i = 1...N_i$ denotes the optimization iteration step number, $j = 1...N_s$ denotes the simulation scenario number, $k = 1...N_m$ denotes the profiling metrics number, $g_i$ is combined evaluation score in the $i$th iteration, $h_i$ is the control parameter set in the $i$th iteration, $s_{j,k}$ is the evaluation score of the $k$th metric in the $j$th scenario, $w_k$ is the weighting coefficient of the $k$th metric, and $m_j$ is the sum of the sample points of the $j$th scenario. Note that for control application in this paper, weighting coefficient $w_k$ is assigned from manual adjustment based on control engineers' experience; while for path planning or other applications, weights can be regressed from human driving data as utilizing Siamese Net with example in Appendix I.

TABLE II: Some typical control profiling metrics used in evaluation score

| Metrics | Normalized Score Computation |
|---|---|
| Lateral error peak | $s = \frac{\max_{n \in 0...N} \|e_{y,n}\|}{e_{y,thold}}$ |
| Lateral error RMS | $s = \sqrt{\sum_{n=0}^{N} \frac{e_{y,n}}{L_w \kappa_{n,des} \dot{x}_{n,des} \Delta t N_f} / N-1}$ |
| Lateral error RMS (harsh) | $s = \sqrt{\sum_{n=0}^{N} \frac{e_{y,n}}{L_w \kappa_{n,des} \dot{x}_{n,des} \Delta t N_f} / N-1}$ only at $n$ where $\kappa_{n,des} > \underline{\kappa}$ |
| Re-plan event | $s = \sqrt{\sum_{n=0}^{N} n_{re} / N-1}$ |

where, $e_{y,thold}$ is the lateral error threshold, $\Delta t$ is the sampling time, $N_f$ is the sample step horizon for normalization, $\underline{\kappa}$ is the curvature lower threshold to define the "harsh" driving zone, and $n_{re}$ is the sample point at which too large control error induces the re-planning trajectory.

*2) Parameter Optimizer (tuner core):* The parameter optimizer, i.e., the core algorithm of auto-tuner, leverages Bayesian Optimization, a machine learning technique, to search an optimal parameter set in the high-dimension parameter space of a control system. It aims to pursue the best performance grade from the "black-box" objective function.

Our Bayesian-based optimizer provides choices of multiple surrogate models (the Bayesian statistical models for approximating the objective function), including Gaussian Process Regression (GPR) [23] and Tree-structured Parzen Estimator (TPE) [24]. Corresponding to a specific surrogate model, the optimizer provides choices of different acquisition functions (the functions to decide where to search at next iteration), i.e, Upper Confidence Bounds (UCB), Expected Improvement (EI) and Max Entropy (ME) for sequential or parallel evaluation executions, continuous or discrete parameters, and independent or relative sampling methods.

Without loss of generality, we explain the Bayesian optimization algorithm with the GPR surrogate model and UCB acquisition function. Let us first define the overall parameter space as $H$ which includes $d$ independent parameters. Then our data set contains: (1) $d$-dimensional training vector $h = [h_1, h_2, ..., h_i] \in \mathbb{R}_d$, in which an individual training point $h_i$ represents the parameter set in the $i$-th iteration and (2) the label vector $g = [g_1, g_2, ..., g_i] \in \mathbb{R}$, in which an individual label $g_i$ represents the evaluation score in the $i$-th iteration.

Once initialized with some labeled training points, the GPR surrogate model (which is a Gaussian process) generates a Bayesian posterior probability distribution to describe a potential score $g^*$ that is normally distributed with mean $\mu(h^*)$ and variance $\sigma_v^2(h^*)$, for any candidate training point $h^*$. Each time we test score at a new point, this distribution is iteratively updated. An acquisition function is employed to predict the next most promising point based on the posterior probability distribution. With the UCB function, the most promising point $h_{i+1}$ suggested by the GPR model is the one with the highest upper-confidence bound:

$$h_{i+1} = \arg\max_{h^* \in H} (\mu(h^*) + \kappa_w \sigma_v(h^*)) \quad (8)$$

$where,$

$$\begin{aligned}
\mu(h^*) &= \mu_0(h^*) + \\
&\quad \Sigma_0(h^*, h)[\Sigma_0(h, h) + \sigma_n^2 I]^{-1}(g - \mu_0(h)) \\
\sigma_v^2(h^*) &= \Sigma_0(h^*, h^*) - \\
&\quad \Sigma_0(h^*, h)[\Sigma_0(h, h) + \sigma_n^2 I]^{-1} \Sigma_0(h, h^*)
\end{aligned} \quad (9)$$

where, $h^*$ is the predicted new training point and $h$ is the tested training vector (parameter sets), $\sigma_n^2$ is the variance of noise, $I$ is the identity matrix, $g$ is the tested label vector (evaluation scores), $\Sigma_0$ is a kernel (covariance function) and $\mu_o$ is a mean function [25]. The weighting factor $\kappa_w$ is selected to balance the efforts between exploration and exploitation. Higher $\kappa_w$ pushes the tuner to explore more potential good points with high uncertainty; while lower $\kappa_w$ drives the tuner to exploit the knowledge about the current best-score point found so far. Thus, the new sample point $h_{i+1}$ is actually the predicted maximum of the acquisition function (8) which combines the mean and variance functions of the posterior probability distribution described by (9).

In summary, in the $i$-th iteration, the tuner core distributes out a suggested optimal parameter set $h_i$ to simulation, and collects back a grading score $g_i$ from tuner grader. Then the new pair $[h_i, g_i]$ is appended to the tracked "parameter set - score" pairs history, which reduces the uncertainties of the predictive posterior distribution in the parameter space $H$. Based on the historical results, the tuner core utilizes the Bayesian surrogate model and acquisition function to suggest a next promising parameter set $h_{i+1}$ that attempts to further

TABLE III: Parameter optimizer core algorithm features

| Surrogate Model | **GPR** | **TPE** |
|---|---|---|
| Acquisition Function | UCB/EI/ME | EI |
| Evaluation Execution | Sequential | Sequential or Parallel |
| Time Efficiency* | ∼120 sec/iteration | ∼30 sec/iteration |
| Training Environment | Cloud cluster with 130 CPUs of i7 processors clocked at 2.6 GHz | |

* Average time estimated with control testing scenario set, including 7 typical scenarios. Parallel mode is used for TPE.

optimize the score and reduce the prediction uncertainties. Table III shows the main model setting features already tested in our parameter optimizer. It is worth noting that the TPE model particularly supports the distributed parallel evaluation process that significantly reduces our optimization time.

## IV. EXPERIMENT AND SIMULATION RESULTS

To validate and expand the application of the parameter auto-tuner in autonomous vehicle control, we present both numerical simulation results and real vehicle road-test results on Apollo Autonomous Driving Platform [1].

For simulation, the auto-tuner (tuner core with simulation environment) is performed under the environment shown in Table III. For road-test, the control module has been verified on two different vehicles, as shown in Fig. 7a-7b (the latter one has only outer-loop control as it is primarily for low-speed education purpose).

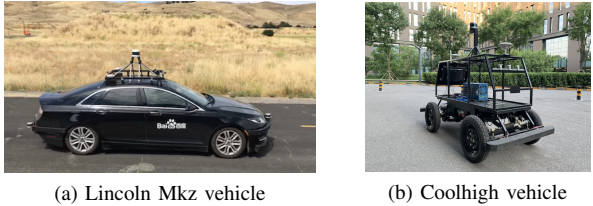

(a) Lincoln Mkz vehicle     (b) Coolhigh vehicle

Fig. 7: Experimental vehicles in Apollo Platform

### A. Inner-Loop Control: Auto-tune Validation by Experiments

As introduced in Section III-A.2, inner-loop control is a relatively independent control component, and hence, can be a good demonstration of the control parameter auto-tune. To highlight the performance improvement brought by the well-tuned MRAC control, we have designed seven typical "simulation scenarios" that contain intense and sharp steering actions (for both the left/right turn and U-turn), various "serpentine-shape" successive side-pass, and regular straight driving.

With the tunable parameters $T_{ref}$ and $p$, the inner-loop control has been auto-tuned from the simulation-based Auto-tuner. Fig. 8a presents the MRAC control parameter auto-tuned results, with both GPR and TPE surrogate models and multiple tuning settings (iteration steps, etc.). Fig. 8b presents the optimization process with highlighted evaluation scores through the entire tunable parameters' space.

The result shows that all the tuned parameters corresponding to multiple settings converge to a global optimal region around the original manual-tuned parameters. This reveals that, with the representative scenarios and comprehensive weighted scores, the auto-tune control parameters are very close to the manual-tuned ones, which can be treated as a good benchmark for auto-tuner optimality validation.

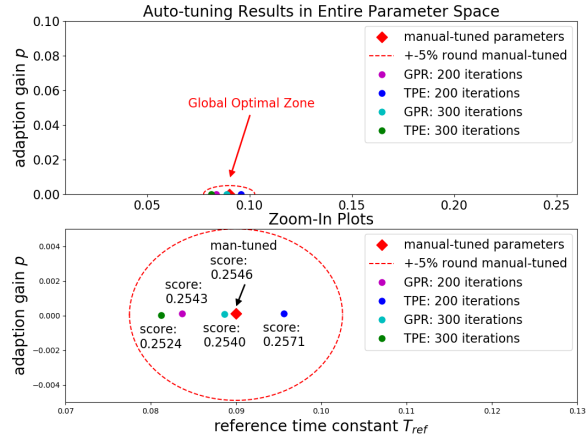

(a) Auto-tune results

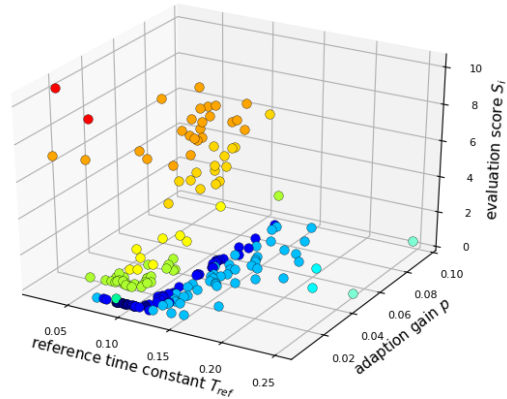

(b) Auto-tune sample points (TPE model, 200 iterations)

Fig. 8: Inner-loop control auto-tune results

Furthermore, the inner-loop MRAC control with tuned parameters has been validated by the road-test on autonomous Lincoln Mkz vehicles. With successive side-pass scenario as typical case, the vehicle always starts from similar positions shown in Fig. 9a. At the mid-way, Fig. 9b shows visible lateral offset from the planning trajectory due to the steer actuation dynamics, which eventually results in the "re-plan" decision in experiments; while Fig. 9c shows much better trajectory-tracking, by taking advantage of MRAC control.

Table IV presents the remarkable lateral control performance improvement from the MRAC control with well-tuned parameters, on successive side-pass experiments. The experimental results fully validate that the auto-tuned control parameters well match the realistic optimal area with significant control performance improvement.

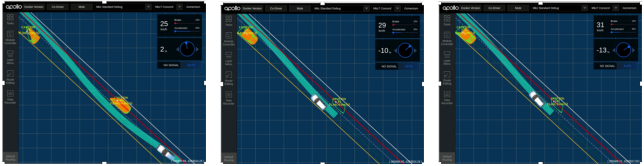

(a) Start side-pass  (b) Midway: no MRAC  (c) Midway: MRAC

Fig. 9: Road-Test: successive side-pass with MRAC on/off

TABLE IV: Control performance with auto-tuned parameters

| Control Performance | Lateral Error Peak | Lateral Error RMS (hash) |
|---|---|---|
| Without MRAC | 111.66% | 39.31% |
| With MRAC | 44.49% | 24.01% |
| Improvement | **+60.16%** | **+38.92%** |

### B. Outer-Loop Control: Auto-tune Expansion

Based on the inner-loop control auto-tune performance validation, we have expanded the application of auto-tuner to a wider domain of complete closed-loop controls, with higher dimensions of (up to 11-dimension) parameters' space.

Fig. 10 presents partial lateral control parameter auto-tuned results (only two most important parameters $Q_1$ and $Q_3$ are presented), with multiple tuning conditions (parameters' dimensions, iteration steps). It is worth noting that, because of the limited-searching scope of our manual-tune process, the auto-tuner provides us some optimized parameters apart from the manual-tune ones, which produces better control performance than manual-tune in simulations.

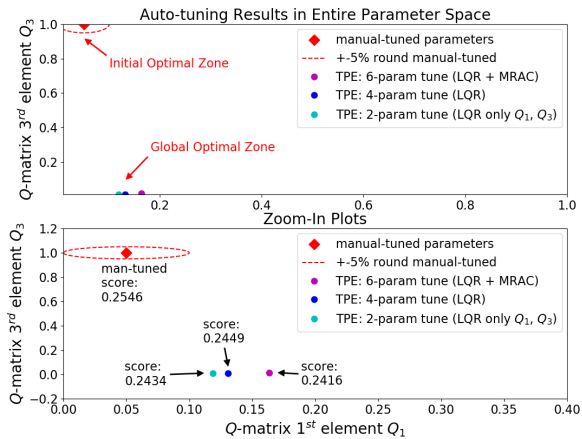

Fig. 10: Lateral control auto-tuned results

Fig. 11a - 11f shows the lateral and longitudinal control performance improvement gained from the auto-tuned control parameters compared with the manual-tuned parameters. The performance improvements with the auto-tuned results will be experimentally verified in road-test in the future.

The closed-loop control parameter auto-tuner not only leads to better control performance, but also improves the time-efficiency of parameter tuning. Table V shows the auto-tuner accomplishes the complete control (11 parameters)

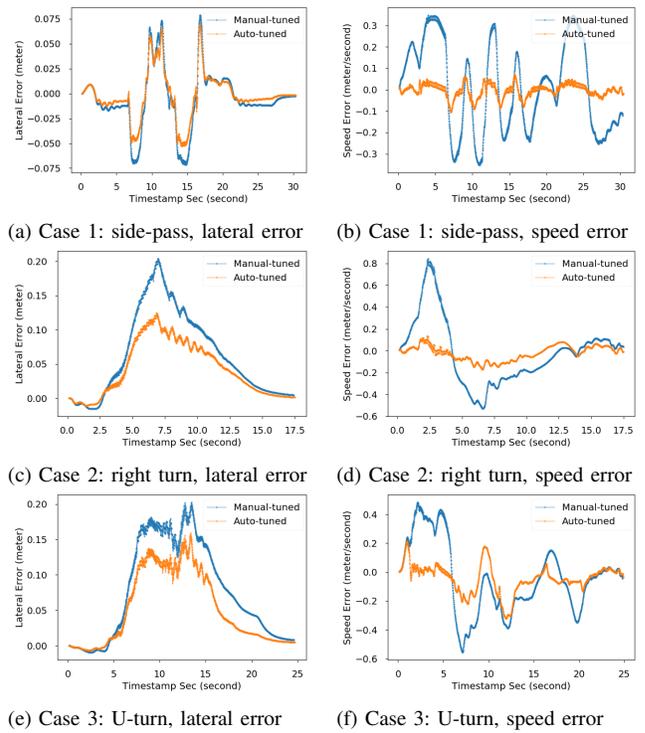

(a) Case 1: side-pass, lateral error  (b) Case 1: side-pass, speed error
(c) Case 2: right turn, lateral error  (d) Case 2: right turn, speed error
(e) Case 3: U-turn, lateral error  (f) Case 3: U-turn, speed error

Fig. 11: Performance comparison of complete control with manual-tune and auto-tuned parameters (11-dimension)

TABLE V: Time efficiencies of auto-tune and manual-tune

| Auto-tuner Setup | Iterations | Tune Time |
|---|---|---|
| MRAC (2-Param) | $\sim 200$ steps | $\sim 1 - 2$ hrs |
| Lateral (6-Param) | $\sim 400$ steps | $\sim 3 - 4$ hrs |
| Complete (11-Param) | $\sim 800$ steps | $\sim 6 - 7$ hrs |
| Manual-tune Setup Complete (11-Param) | $\sim 5$ full days | overall 120hrs |

tuning by about 18 times faster than regular manual-tune. In addition, the control robustness of the auto-turned parameters is verified by the successful application in thousands of simulation scenarios (229 of which are open to public) covering abundant environments, routes and initial states, without loss of stability.

## V. CONCLUSION

This paper presents a framework that utilizes three different learning-based approaches to make the autonomous driving control tune-free for diverse vehicles. The framework includes 1) a learning-based dynamic model to enable control-in-the-loop simulation; 2) a learning-based open-loop mapping to generate feedforward control parameters; and 3) a Bayesian-based auto-tuner to automatically tune closed-loop control parameters in simulation, with 16x, 20x and 18x tuning-efficiency increase respectively compared with manual-tune methods, and remove engineer's labor out of tuning loop. One drawback of current auto-tuner is that for control parameters' auto-tuning, as shown in Section III-B.1, the weighting coefficients used in tuner grader have to be tuned manually (through once fixed, these weights are applicable to different vehicle models). In the future, the automatic training of weight coefficients will be investigated and we may also extend this auto-tune framework to other autonomous driving modules including planning module and perception module (for object tracking function tuning).

## APPENDIX I
### SIAMESE NET METHOD FOR WEIGHT REGRESSION

The Siamese Net is designed for weight regression in tuner grader for planning parameters' tuning applications, with main objective to adjust the weight coefficients to minimize the difference between human driving data and generated trajectories.

$$Loss = \sum_{j=0}^{N_s} |g(w; \hat{\xi}_j) - g(w; \xi_j) + \varepsilon|_+ \qquad (10)$$

where, $\hat{\xi}$ and $\xi$ are human driving and generated planning trajectory respectively, $|\cdot|_+$ is the maximum between $\cdot$ and $0$, $j = 1...N_s$ is the testing scenario number, $\varepsilon$ is the marginal factor which is a small positive constant. We tried to solve the optimization problem to minimize the objective function defined in (10). The marginal constant $\varepsilon$ is necessary added to avoid getting all zero weights. Also, the relatively small value of $\varepsilon$ can omit the affect of the sampled trajectories with costs much larger than the ground-truth trajectory cost, in which situation, $|g(w; \hat{s}_j) - g(w; s_j) + \varepsilon|_+ = 0$.

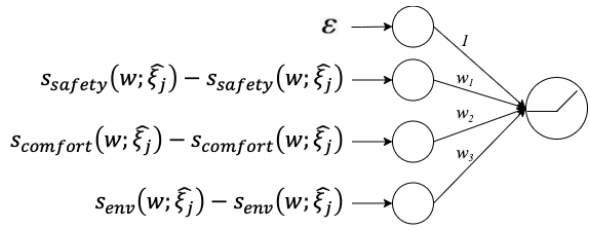

Fig. 12: An example of using Siamese net weight regression in planning module